\documentclass[10pt,twocolumn,letterpaper]{article}

\usepackage[pagenumbers]{cvpr} 

\usepackage{graphicx}
\usepackage{amsmath}
\usepackage{amssymb}
\usepackage{booktabs}
\usepackage{multirow}
\usepackage{multicol}
\usepackage{float}

\usepackage[pagebackref,breaklinks,colorlinks]{hyperref}

\usepackage[capitalize]{cleveref}
\crefname{section}{Sec.}{Secs.}
\Crefname{section}{Section}{Sections}
\Crefname{table}{Table}{Tables}
\crefname{table}{Tab.}{Tabs.}

\usepackage{color}
\definecolor{green}{rgb}{0, 1.0, 0}
\definecolor{blue}{rgb}{0, 0.0, 1.0}
\definecolor{red}{rgb}{1.0, 0.0, 0.0}
\newcommand{\ncj}[1]{{\color{black}#1}}
\newcommand{\rz}[1]{{\color{black}#1}}

\usepackage{verbatim}

\def\cvprPaperID{1743} 
\def\confName{CVPR}
\def\confYear{2022}

\begin{document}

\title{RIM-Net: Recursive Implicit Fields for \\ Unsupervised Learning of Hierarchical Shape Structures}

\author{
Chengjie Niu$^1$~~~~~~~~
Manyi Li$^2$~~~~~~~~
Kai Xu$^1$\thanks{Corresponding author: kevin.kai.xu@gmail.com}~~~~~~~~
Hao Zhang$^3$~~~~~~~~
\smallskip
\\
$^1$National University of Defense Technology~~~~
$^2$Shandong University~~~~
$^3$Simon Fraser University
}

\maketitle

\begin{abstract}


We introduce RIM-Net, a neural network which learns {\em recursive implicit fields\/} for {\em unsupervised\/} inference of hierarchical shape structures. Our network recursively decomposes an input 3D shape into two parts, resulting in a binary tree hierarchy. Each level of the tree corresponds to an assembly of shape parts, represented as implicit functions, to reconstruct the input shape. At each node of the tree, simultaneous feature decoding and shape decomposition are carried out by their respective feature and part decoders, with weight sharing across the same hierarchy level. As an implicit field decoder, the part decoder is designed to decompose a sub-shape, via a {\em two-way branched reconstruction\/}, where each branch predicts a set of parameters defining a Gaussian to serve as a local point distribution for shape reconstruction. With reconstruction losses accounted for at each hierarchy level and a decomposition loss at each node, our network training does not require any ground-truth segmentations, let alone hierarchies.
Through extensive experiments and comparisons to state-of-the-art alternatives, we demonstrate the quality, consistency, and interpretability of hierarchical structural inference by RIM-Net. 
\end{abstract}

\section{Introduction}
\label{sec:Intro}

``What emerges is a {\em multileveled hierarchical structure of parts and wholes\/}, each of which has a representation of holistic
properties as well as component structure.''
\vspace{-5pt}
\begin{flushright}
--- Stephen E. Palmar~\cite{palmer1977}
\end{flushright}

\begin{figure}
    \centering
    \includegraphics[width=1\linewidth]{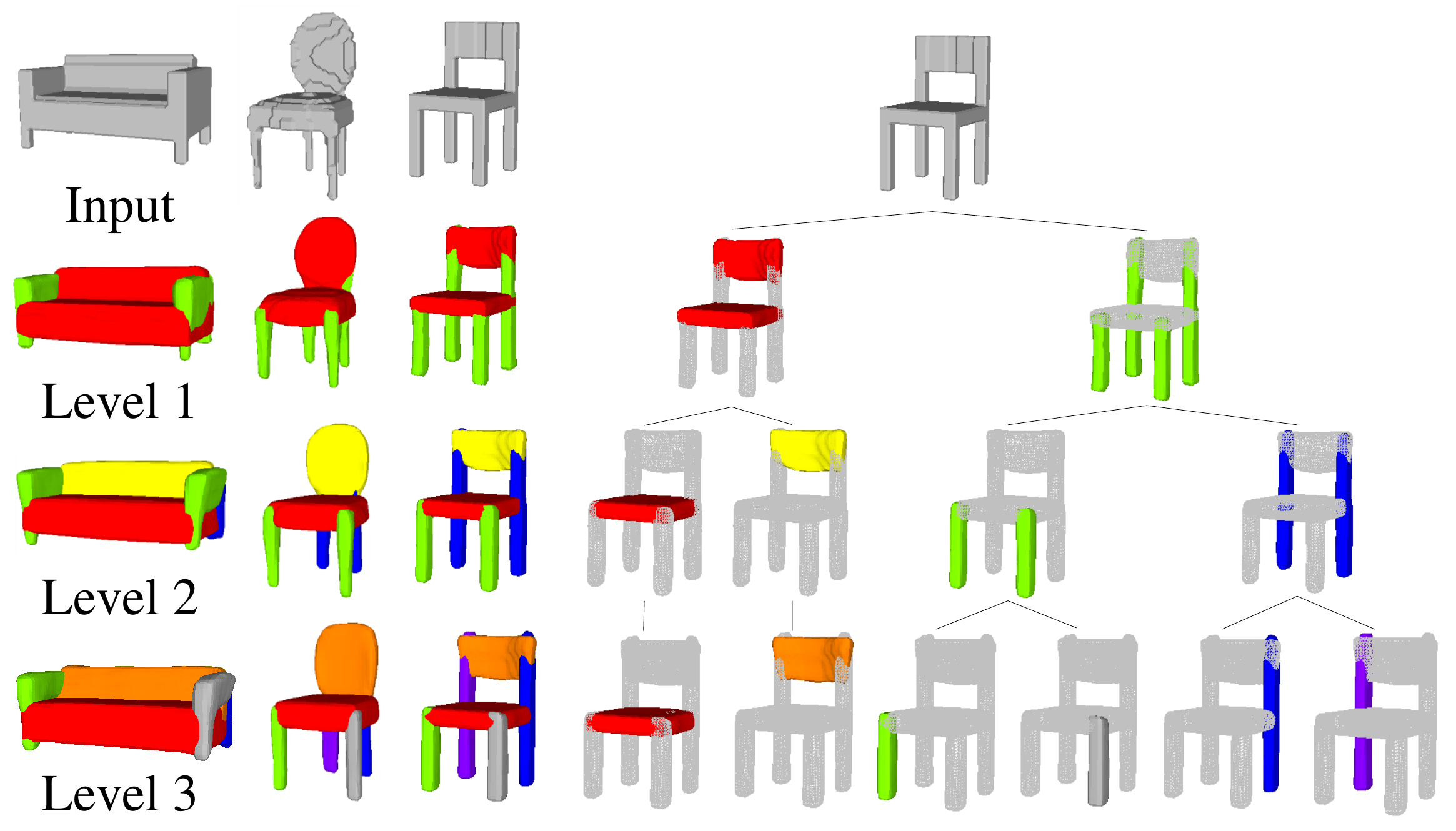}
    \caption{Hierarchical shape structures predicted by RIM-Net, which was trained on sofas and chairs together. Parts predicted by the same branch of the network are assigned the same color. We observe structural consistencies across multiple levels, with interpretability and shape semantics revealed by the hierarchies.}
    \vspace{-3pt}
    \label{fig:teaser}
\end{figure}

The recent emergence of neural implicit representations for 3D shapes~\cite{chen2019learning,mescheder2019occupancy,park2019deepsdf} has stimulated much follow-up. One line of research is motivated by the question of whether self-supervision using the reconstruction loss would allow a neural network to learn a {\em structured\/} implicit representation that reveals semantic shape parts.
Chen et al.~\cite{chen2019bae} gave a positive answer: by adding a branching layer to the original IM-Net~\cite{chen2019learning} for learning holistic implicit fields, the resulting network can be trained to provide a consistent co-segmentation over a large set of shapes. However,  such a branched autoencoder, or BAE-Net, can only return a coarse segmentation, especially amid structural variations in the set, and the parts obtained are all unorganized.

Cognitive psychology studies have long suggested that human shape perception is based on a {\em hierarchical\/} organization of shape parts which encode both per-part properties and part-to-part, as well as part-in-whole, relations~\cite{palmer1977,hoffman1984,thompson1992}. Such a hierarchy is not only a better reflection of object functionality~\cite{carlson1999}, but also a more granular representation that can better capture the structural variability in a diverse shape collection, from coarse organization to finer-level structures.
The key question then is whether contemporary neural implicit models are capable of learning structural hierarchies for 3D shapes, {\em without supervision\/}.

\begin{figure*}
    \centering
    \begin{subfigure}{0.36\textwidth}
    \centering
 \vspace{0.15in}
    \includegraphics[width= 0.8\textwidth]{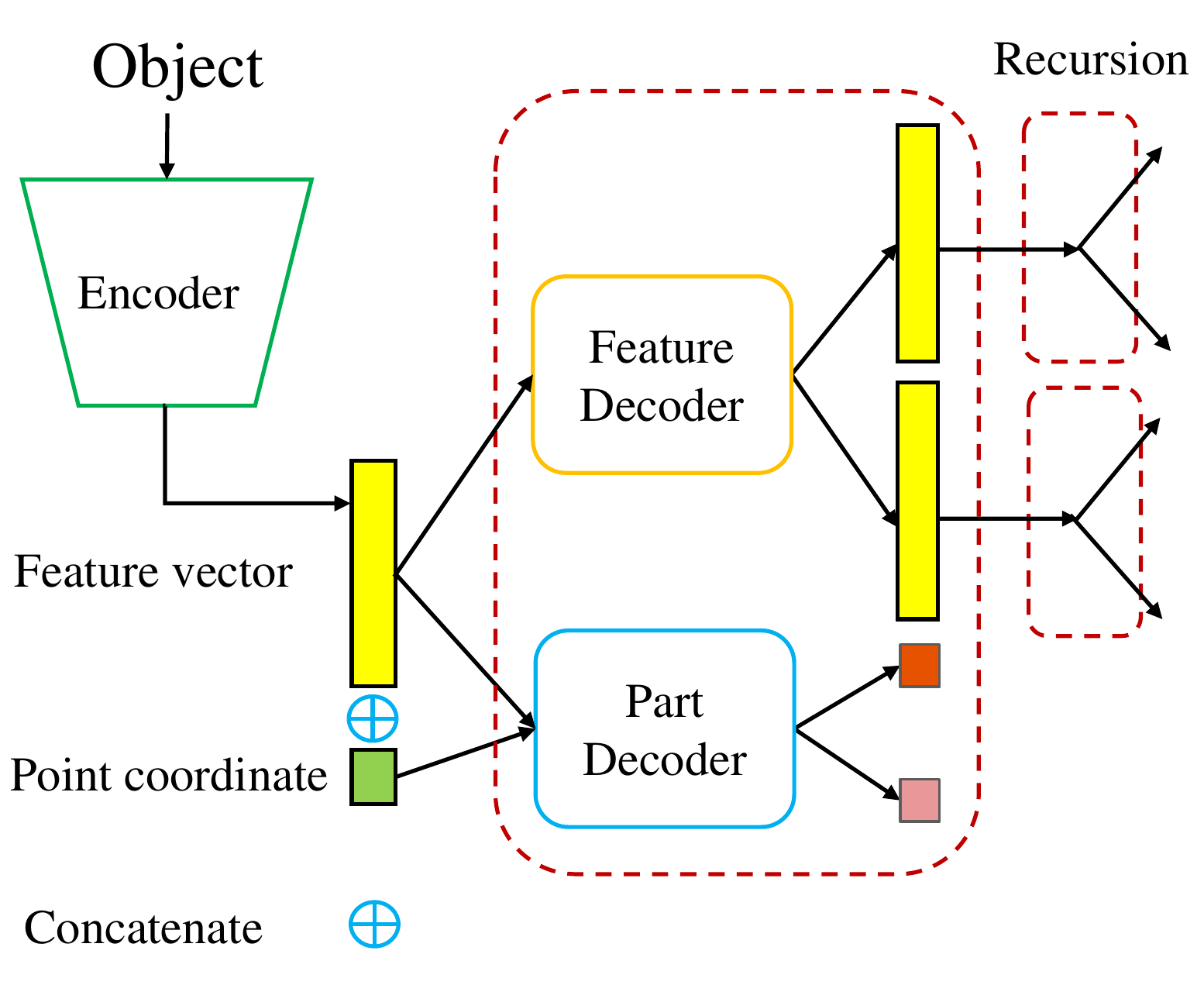}
    \vspace{0.07in}
    \caption{Overall model architecture.}
    \label{fig:arch-a}
    \end{subfigure}
    \begin{subfigure}{0.31\textwidth}
    \centering
    \includegraphics[width= 0.8\textwidth]{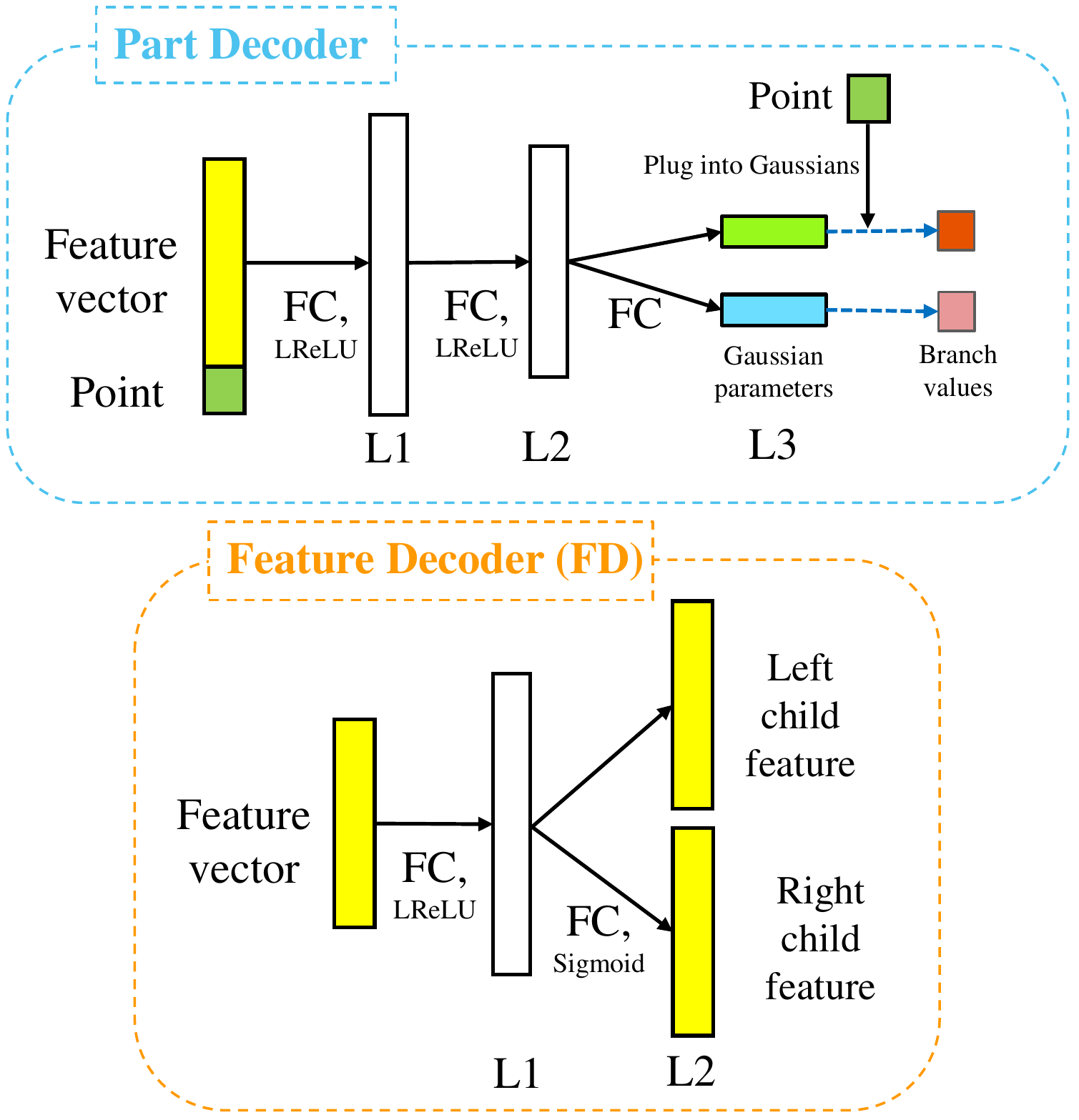}
    \vspace{0.05in}
    \caption{Two decoders at each node.}
    \label{fig:arch-b}
    \end{subfigure}
    \begin{subfigure}{0.31\textwidth}
    \centering
    \vspace{0.00in}
    \includegraphics[width= 0.8\textwidth]{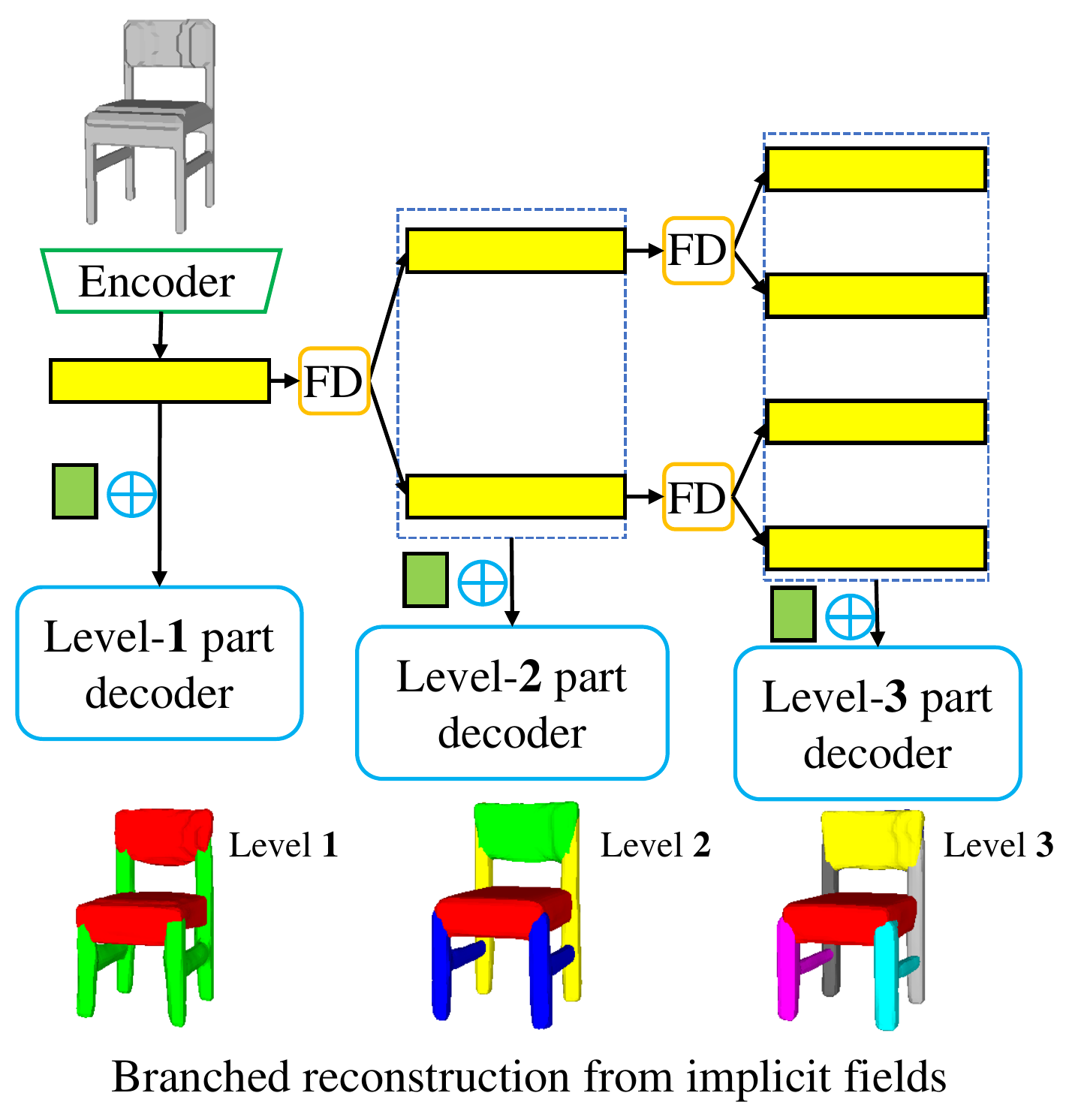}
    \vspace{0.025in}
    \caption{Three-level RIM-Net with inference.}
    \label{fig:arch-c}
    \end{subfigure}
    \caption{Overall network architecture of RIM-Net (a), the two decoder modules (b), and a three-level RIM-Net at work on a 3D chair model (c). After encoding the input 3D shape into a feature vector, the network operates in the feature space with simultaneous feature decoding (FD) and part decomposition (via the part decoder) at each node and recursively down the hierarchy. The part decoder predicts an implicit field via a two-way branched reconstruction, where each branch predicts a set of Gaussian parameters to indicate one part.}
    \vspace{-3pt}
    \label{fig:arch}
\end{figure*}

In this paper, we introduce RIM-Net, a neural network which learns {\em recursive implicit fields\/} for {\em unsupervised\/} inference of hierarchical shape structures. The network recursively decomposes an input 3D shape into two parts, resulting in a binary tree hierarchy. Each level of the tree corresponds to an assembly of shape parts, represented as implicit functions, to reconstruct the input shape.
Our network employs a reconstruction loss, like most neural implicit models~\cite{chen2019learning,chen2019bae,mescheder2019occupancy,park2019deepsdf}. In our setting, the loss is applied at {\em each\/} level of the structure hierarchy and summed up.
Furthermore, we add a {\em decomposition loss\/}, defined at each tree node, to ensure that the shape it represents is a union of the two shapes corresponding to its two child nodes.

\cref{fig:arch} shows the architecture of RIM-Net, which takes as input a 3D voxel shape and first encodes it into a feature vector using a conv-net. The network then operates in the feature space recursively through two modules:
\begin{itemize}
\item A {\em feature decoder\/} which inputs a feature code and produces two child feature vectors to infer shape parts.
\item A {\em part decoder\/} which takes a feature code $c$ and a 3D point $p$ as input and eventually outputs an occupancy value for $p$ with respect to the shape part represented by $c$. The part decoder is designed to decompose this shape part into two sub-parts, by way of a  {\em branched reconstruction\/}, as in BAE-Net~\cite{chen2019bae}. A key difference to a two-way BAE-Net is that each branch predicts a set of parameters defining a {\em Gaussian\/}, instead of a point-wise occupancy. These Gaussians serve as {\em local point distributions\/} for shape reconstruction.
\end{itemize}

Our network is unsupervised as it does not require any ground-truth segmentations, let alone hierarchies. Network weights are shared between all part decoders, respectively all feature decoders, at the same level of the hierarchy. Between different levels, the weights are different.

\begin{figure}
    \centering
    \includegraphics[width=0.9\linewidth]{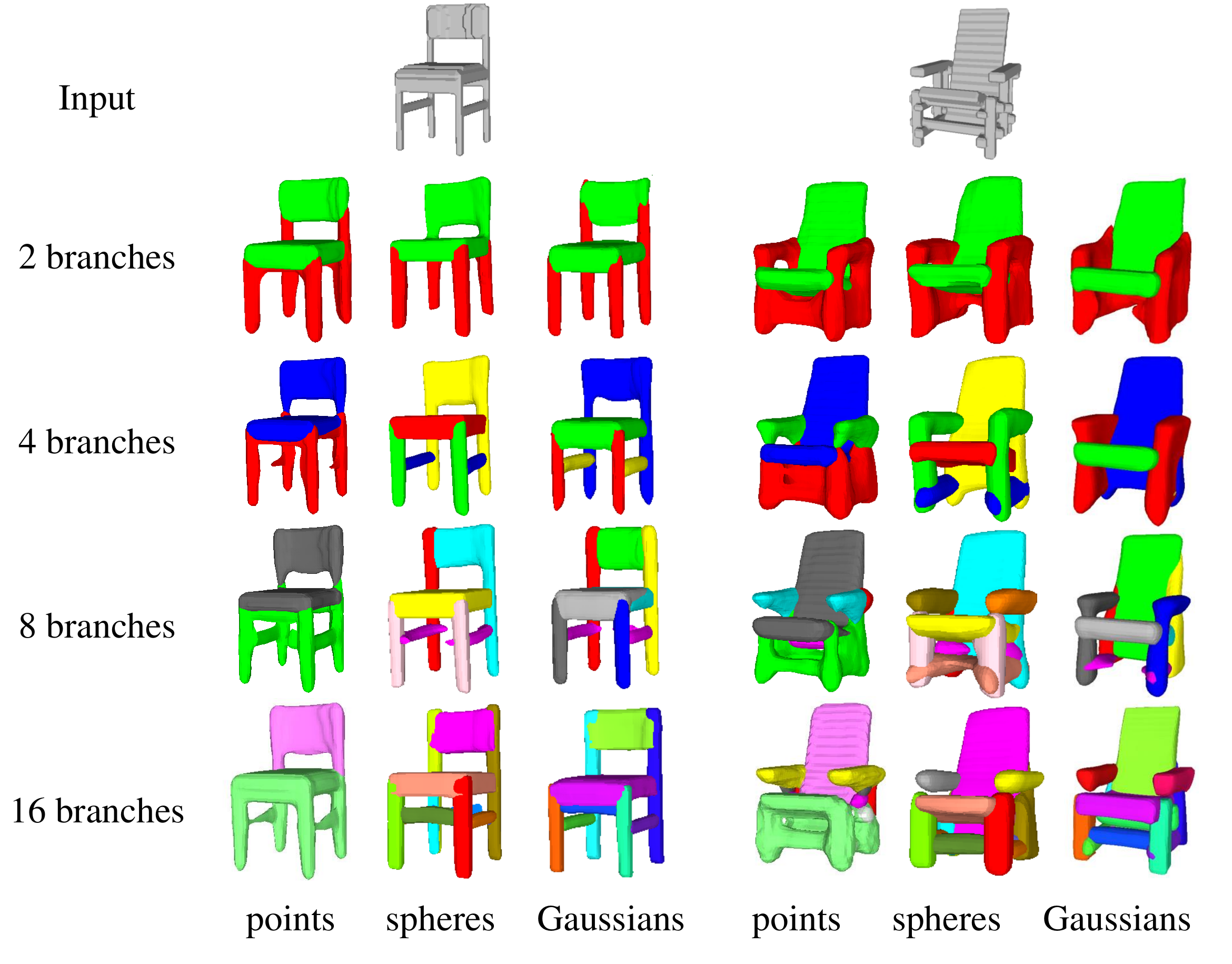}
    \caption{Comparing {\em single-level\/} (i.e., no hierarchy) branched neural implicit reconstruction using different local point distributions. Higher degrees of freedom, i.e., from points~\cite{chen2019bae} to spheres, then to Gaussians, improve reconstruction and part inference.}
    \label{fig:point_vs_gaussian}
\end{figure}

RIM-Net is the first unsupervised, hierarchical neural implicit model which learns recursive 3D shape decomposition. The idea of using per-point Gaussian inference in the part decoder is also novel. Compared to predicting only a single value per branch, as in BAE-Net~\cite{chen2019bae}, the Gaussians offer greater degrees of freedom for the part decoders to adapt to the geometric varieties among parts that can be reconstructed  through each branch. This leads to improved reconstruction and segmentation, as shown in \cref{fig:point_vs_gaussian}.

We evaluate RIM-Net and compare to state-of-the-art methods for hierarchical shape abstraction~\cite{tulsiani2017learning,yang2021unsupervised,paschalidou2019superquadrics,sun2019learning,chen2019bae} and structure inference~\cite{chen2019bae,paschalidou2020learning}. 
Experimented tasks include 3D shape autoencoding, co-segmentation, and single-view 3D reconstruction. We also conduct ablation studies to validate our network designs and training strategies.

\section{Related Work}
\label{sec:related}

\paragraph{Semantic segmentation.}
%
%
Most methods on learning semantic shape segmentation are supervised, e.g.,~\cite{qi2017pointnet, qi2017pointnet++, zhao20193d,kalogerakis20173d}. As a representative weakly supervised approach,
Tags2Parts \cite{muralikrishnan2018tags2parts} obtains semantic part annotations from weak shape-level tags through a deep neural network, which is trained to classify the shape as having or lacking a part.
AdaCoSeg \cite{zhu2020adacoseg} learns adaptive co-segmentation over a set of shapes using a group consistency loss.
RIM-Net is the first unsupervised method for {\em hierarchical\/} structure inference to produce fine-grained semantic segmentations.

\vspace{-5pt}

\paragraph{Shape abstraction.}
%
Shape abstraction aims to approximate a shape using a compact set of simple geometric primitives.
Due to their simplicity, cuboids have been widely adopted~\cite{tulsiani2017learning,zou20173d,li2017grass, niu2018im2struct,sun2019learning,yang2021unsupervised}.
%
%
Tulsiani et al.~\cite{tulsiani2017learning} designed a deep convolutional neural network to predict the shapes of volumetric primitives (VP) and transformation parameters to assemble a given shape; they only considered cuboids as their VPs. %
In 3D-PRNN, Zou et al.~\cite{zou20173d} developed a recurrent neural network to generate a sequence of cuboids to construct a 3D shape.
More recently, Yang and Chen~\cite{yang2021unsupervised} developed an unsupervised learning method to map a point cloud into a compact cuboid representation by jointly solving the cuboid abstraction (CA) and shape co-segmentation problems.
Paschalidou et al. \cite{paschalidou2019superquadrics} abstracted 3D objects using superquadrics (SQ), which encompass cubes, cylinders, spheres, and ellipsoids, etc. Such a representation is more general and leads to more accurate shape abstractions and faster optimization with continuous parametrization.

While all of the above shape abstraction methods are unsupervised, they do not infer structure hierarchies. The most important distinction to our work is that our goal is not to abstract, but to reconstruct, using a set of implicit shapes.

\vspace{-5pt}

\paragraph{Learning hierarchies.}
%
Notable supervised methods for hierarchical structural analyses include GRASS~\cite{li2017grass} and PartNet~\cite{yu2019partnet}, both employing recursive neural networks (RvNNs) to produce trees, and StructureNet~\cite{mo2019structurenet}, which generalizes trees to hierarchical graph organizations. While RIM-Net also consists of a recursively constructed hierarchy of networks, it is not an RvNN, as the latter implies weight sharing between networks at all tree nodes.

Under the category of unsupervised methods, early work by van Kaick et al.~\cite{conshier_sig13} introduced the co-hierarchical analysis problem over a shape collection and solved it based on multi-instance clustering.
%
%
The hierarchical (cuboid) abstraction (HA) method of Sun et al.~\cite{sun2019learning} infers a three-level hierarchy for an input shape by {\em adaptively\/} selecting from a hierarchical cuboid representation that is shared by all objects in a given class, where the number of cuboids per level is pre-determined.
Most recently, the work of Paschalidou et al.~\cite{paschalidou2020learning} performs hierarchical structure recovery from a single-view image, without part supervision. It recursively extracts superquadric surfaces from the captured shape to construct an unbalanced binary structure hierarchy, as guided by a part prior which clusters points based on part centroids akin to $k$-means clustering.

In contrast, part inference by RIM-Net is not limited by any primitive type, while such limitations can comprise a method's ability to handle unusual part geometries and structure variabilities. Our network is more robust against these issues, owing to the versatility of implicit field representations, as well as the many degrees of freedoms afforded by the local Gaussian models in our part decoder.

\vspace{-5pt}

\paragraph{Structured neural implicits.}
The work by Genova et al. \cite{genova2019learning} can also be regarded as an abstraction method, where the template shapes are defined by a set of local implicit functions which model scaled and axis-aligned 3D Gaussians. Their use of Gaussians provided inspirations to our design of the part decoder. However, instead of employing a {\em fixed\/} number of Gaussians as a template to abstract a shape collection, our network learns {\em per-point Gaussians\/} for faithful shape reconstruction.
Another closely related work which also proved inspiring was BAE-Net~\cite{chen2019bae}, a branched autoencoder for unsupervised or few-shot shape co-segmentation. The part decoder of our network resembles a two-branch BAE-Net, except for the use of per-point Gaussians.
%

In Section~\ref{sec:Results}, we compare our work to the most relevant methods mentioned above, 
including VP~\cite{tulsiani2017learning}, CA~\cite{yang2021unsupervised}, SQ~\cite{paschalidou2019superquadrics}, HA~\cite{sun2019learning}, and BAE-Net~\cite{chen2019bae}. 
\section{Methods}
\label{sec:Methods}


Given a volume as input, we propose a novel network to predict the structure tree where each node corresponds to an implicit field representing a part in different granularity.
%
%
\cref{fig:arch-a} illustrates our network architecture. The encoder is a 3D convolutional neural network~\cite{maturana2015voxnet} for voxels. It maps the input to a feature code of the entire shape, as the root of the structure tree. The decoder is composed of two modules: the feature decoder and the part decoder. The feature decoder takes a feature code as input and outputs two child feature vectors. The part decoder learns the mapping from each node feature to two child parts in 3D, represented as implicit fields. It takes the parent feature and a point coordinate as input, and predicts two values indicating the probability of the input point to be inside each of the two child parts respectively. Therefore, the two modules of the decoder recursively build the structure tree and decode the nodes to parts, forming the recursive implicit fields. The deeper the level, the finer the structure.

Note that the feature decoder and part decoder has shared weights within each level of the tree, but different weights between the levels. Therefore, to train the network for a $N$-level structure tree, we need a total of $N$ different part decoders and $N-1$ different feature decoders.

\subsection{Part Decoder}

We represent each part with a per-point Gaussian which can be regarded as a local point distribution~\cite{genova2019learning}. We devise a novel point-based decoder to recover the per-point Gaussian.
\rz{It takes the concatenation of a feature code $c$ and a point coordinate $p$ as input, and outputs \rz{two sets of Gaussian parameters, one set per child part.} The input point $p$ is plugged into each Gaussian to compute the occupancy probability of $p$ with respect to the corresponding part. Specifically, point $p$ is inside a part if the corresponding probability is larger than a pre-defined threshold. This per-point Gaussian-based decoding} generates more smooth and detailed shapes than other popular decoder networks~\cite{chen2019learning, chen2019bae, genova2019learning}.
	
The part decoder is implemented with a 3-layer MLP as shown in \cref{fig:arch-b}. The first two fully-connected layers are each followed by a LeakyReLU function. The output of the last layer is split into two branches, each being a scaled, anisotropic 3D Gaussian $\theta_i \in \mathbb{R}^7$ consisting of a scaling factor $s_i$, a center point $ \bold{c}_i \in \mathbb{R}^3 $, and a per-axis radii $\bold{r}_i \in \mathbb{R}^3 $. The probability of the input point ${p} \in \mathbb{R}^3$ w.r.t. each branch is formulated as
\begin{equation}\label{equ:gaussian}
	f(p, \theta_i) = s_i \cdot \exp\left({\sum\limits_{d \in \{1,2,3\}}\frac{-(\bold{c}_{i,d} - {p}_d)^2}{2\bold{r}_{i,d}^2}}\right),
\end{equation}
where $s_i$ is clamped into $(0,1]$. Note $s_i$ cannot take $0$ value to avoid a vanishing gradient.

\subsection{Feature Decoder}

The feature decoder maps one parent feature into two child features. It consists of two fully connected layers, as shown in \cref{fig:arch-b}. The first layer is followed by LeakyReLU and the second by a Sigmoid. The output is then split into the left and right child features. All feature codes are 128D vectors.

\subsection{Network Loss Functions}

Our training set includes the point-value pairs sampled from the ground-truth implicit field of the shapes, as in~\cite{chen2019bae}. The ground-truth value of each sample point is one if it is inside the shape and zero otherwise. No part label or structure information is given in the training set.

\vspace{-5pt}

\rz{
\paragraph{Reconstruction Loss.}
Shape parts obtained at each level of the hierarchy should together, via a union, reconstruct the input shape.} We define $f_{i,j}({p})$ as the predicted probability of point ${p}$ being inside the part corresponding to the  $i$th node at the $j$th level. Then the maximum of the set $\{f_{i,j}({p}), i=1,...2^j\}$ is the probability of point ${p}$ being inside the union of these parts. Therefore, the reconstruction loss of point ${p}$ at level $j$ is defined as
\begin{equation}\label{equ:reconi}
	\begin{aligned}
		L_{recon_j}({p}) = (y_{gt} -  \max(\{f_{i,j}({p})\}) )^2,
		i=1,\ldots,2^j
	\end{aligned}
\end{equation}
where $y_{gt} \in \{0,1\}$ is the ground-truth value for point ${p}$ in the training set. We sum up the losses of all levels to form the total reconstruction loss:
\begin{equation}\label{equ:recon}
	L_{recon}^{(N)}({p}) = \sum_{j=1}^N L_{recon_j}({p}).
\end{equation}

\vspace{-5pt}

\rz{
\paragraph{Decomposition loss.}
In addition to shape reconstruction, we explicitly enforce a decomposition relation between an internal node and its two children. Geometrically, the part represented by an internal node should be the union of its two child parts. This constraint is imposed by a {\em decomposition loss\/}. Specifically,} given a point ${p}$, its probability of belonging the parent node, $f_{i,j}({p})$, should be equal to the maximum probability of belonging one of the two child nodes, i.e., $f_{2i-1,j+1}({p})$ and $f_{2i,j+1}({p})$. So the hierarchy loss for the $i$th node at level $j$ is
\begin{equation}\label{equ:hiei}
L_{hie_{i,j}}({p}) = ( f_{i,j}({p}) - \max(f_{2i-1,j+1}({p}), f_{2i,j+1}({p})))^2.
\end{equation}
The total \rz{decomposition} loss at a point ${p}$ is the sum of the $L_{hie_{i,j}}({p})$ for all internal nodes. For a tree with $N$ levels, we have
\begin{equation}\label{equ:hie}
	L_{hie}^{(N)}({p}) = \sum_{j=1}^{N-1}(\sum_{i=1}^{2^j} L_{hie_{i,j}}({p})).
\end{equation}

Finally, our \rz{full} loss function is formulated as follows:
\begin{equation}\label{equ:loss}
	L^{(N)} = \sum_{{p} \in P}(\alpha L_{recon}^{(N)}({p}) + \beta L_{hie}^{(N)}({p})),
\end{equation}
where $P$ is the set of all the sample points in the training set and $N$ the number of levels of the structure tree. We set $\alpha=1$ and $\beta=10$ during the training process.

\subsection{Training Strategy}
\label{sec:training}

We propose a progressive learning strategy to ensure stable and efficient network training. To infer an $N$-level structure tree, our training process proceeds in three stages:
\begin{itemize}
\item {\em Initial training.}
The first stage uses the loss function $L^{(1)}$ (\cref{equ:loss}) at the first level to bootstrap the network training. Note that the \rz{decomposition} loss is always zero when $N=1$. Hence, this stage only trains the encoder and a part decoder at the first level, with the goal being to generate a shape assembled by two parts.
\item {\em Recursive training.}
During this stage, we progressively train each level of the network. While training the feature decoder and part decoder at level $j$, we fix the weights and bias of the encoder and networks at the previous level, and use the loss $L^{(j)}$ (see \cref{equ:loss}). This process is repeated for levels $j=2,\ldots,N$.
\item {\em Fine-tuning.}
Finally, we fine-tune the whole network to learn finer details with the loss $L_{recon}^{(N)}$ in \cref{equ:recon}.
\end{itemize}

\section{Results and Evaluation}
\label{sec:Results}

\begin{figure*}
    \centering
    \includegraphics[width=.99\textwidth]{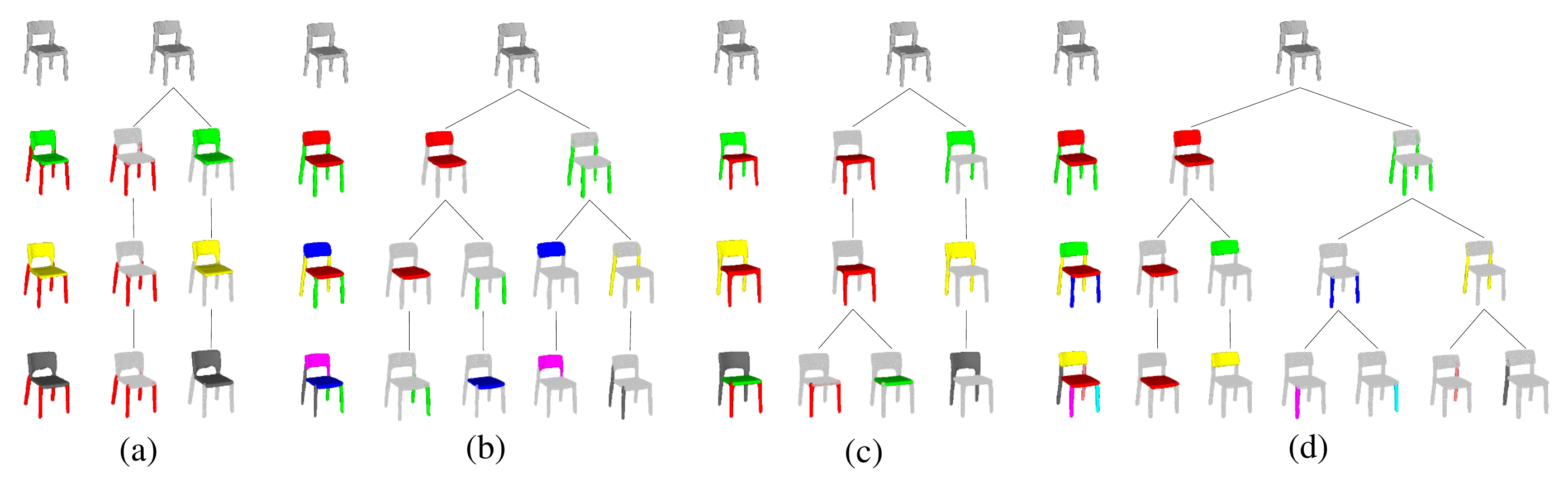}
    \caption{Qualitative results of ablation experiments. 3D hierarchical structure reconstruction with different settings: (a) \rz{without per-point Gaussians}; (b) without \rz{decomposition} loss; (c) without progressive training; (d) RIM-Net.  }
    \label{fig:ablation}
\end{figure*}

We first analyze the novel designs of our method in learning good hierarchical shape structures. We then compare with several baseline methods to demonstrate the superiority of our learned structure. We also evaluate our method with downstream applications.
	
\subsection{Training Details and Metrics}
	
We use the ShapeNet~\cite{chang2015shapenet} subset of Choy et al.~\cite{choy20163d} in our experiments. The data preparation follows~\cite{chen2019bae} with the same train/test split, shape voxelization, as well as the ground-truth point-value samples.
We run our experiments and comparative studies on three object categories: airplanes (2,690), chairs (3,758), and tables (5,271), and train an individual model for each category. During our training, we set the batch size to 1 and run $100$K iterations per stage in the progressive training. The output meshes are extracted using Marching Cubes at the resolution of $64^3$. 
	
We evaluate our RIM-Net and the related approaches from the aspects of structure learning and shape reconstruction. For the structure learning, we use the mean IoU (mIoU) in unsupervised segmentation task to assess the meaningfulness and consistency of the inferred structures, similar to BAE-Net~\cite{chen2019bae}. For reconstruction, we adopt the popular symmetric Chamfer Distance (CD) and IOU for evaluating the overall accuracy and Light Field Distance (LFD) for visual quality.
In particular, CD is computed on 4,096 points uniformly sampled from the reconstructed meshes, and IoU is computed on $32^3$ volume.

\subsection{Ablation Study}

We validate several key designs of our method including the per-point Gaussian-based decoding module, the per-node \rz{decomposition} term in the network loss, and the progressive training strategy.
\cref{fig:ablation} gives a qualitative comparison, with distinct colors indicating different parts.
\ncj{The quantitative results are shown in \cref{tab:ablation}. We see that RIM-Net outperforms other ablated networks in both reconstruction quality and segmentation accuracy.}

\vspace{-5pt}

\paragraph{Per-point Gaussian.}
\rz{
We conduct experiments to evaluate the use of per-point Gaussians by RIM-Net for implicit part reconstruction.
In \cref{fig:ablation}(a), we show a structure hierarchy predicted by the same network architecture as RIM-Net, except that the part decoder only outputs a scalar probability value per point. This is contrasted to the result in (d) by RIM-Net, demonstrating that the per-point Gaussians help  obtain finer structures and better visual quality. For example, the decomposition of chair back and seat cannot be split in (a) and there is no more segmentation at level 2 and 3.

For a further demonstration to isolate advantages of using the Gaussians, we remove the hierarchy and only keep the first-level part decoder in RIM-Net, operating on the root feature vector output from the 3D conv-net encoder. The compared architectures are the same except they output different local point distributions for shape reconstruction: points (i.e., single value per point) \cite{chen2019bae} and spheres (consisting of a scaling factor, a center point, and a radii). We tried different numbers of branches in the part decoder, two and up, to provide a more general picture. \cref{fig:point_vs_gaussian} shows that with higher degrees of freedom provided by per-point Gaussians, the network tends to improve on both reconstruction and part inference: it can obtain fine-grained segmentation of the shape, the more part branches, the finer the division, while the method with points fails to segment shape finer.}


\vspace{-5pt}

\paragraph{Decomposition loss.}
In learning hierarchical structures via reconstruction, we use a \rz{decomposition} loss to constrain the parent-children decomposition relation in the hierarchy. We train an ablated version of our RIM-Net without decomposition loss, keeping the other settings unchanged. In \cref{fig:ablation}(b), we see the ablated model fails to maintain the parent-children relation in various levels and the decomposition cannot go finer compared with our full method. This verifies that the reconstruction loss only is not enough to learn a good hierarchical structure.

\vspace{-5pt}

\paragraph{Progressive training.}
Our method progressively trains each level and then finetunes the full network. Here, we come up with a baseline without the progressive training strategy. It trains the full network jointly from scratch. The number of iterations is the same as that for training one level in our progressive method. As shown in \cref{fig:ablation}(c), our method is unable to achieve a fine-grained and reasonable segmentation without the progressive training.

\begin{table}[]
 \centering
\footnotesize
\setlength\tabcolsep{2.5pt}
\renewcommand{\arraystretch}{0.9}
\begin{tabular}{cc|c|c|c|c}
\toprule
\multicolumn{2}{c|}{}                                                                                    & \begin{tabular}[c]{@{}c@{}}w/o per-point \\ Gaussian\end{tabular} & \begin{tabular}[c]{@{}c@{}}w/o deco.\\  loss\end{tabular} & \begin{tabular}[c]{@{}c@{}}w/o prog.\\  training\end{tabular} & RIM-Net            \\ \hline
\multicolumn{1}{c|}{\multirow{4}{*}{CD}}                                                      & airplane & 0.2336                                                            & \textbf{0.1881}                                                   & 0.3403                                                              & 0.2228          \\
\multicolumn{1}{c|}{}                                                                         & chair    & 0.4588                                                            & 0.4770                                                            & 0.8620                                                              & \textbf{0.4125} \\
\multicolumn{1}{c|}{}                                                                         & table    & 0.7804                                                            & 0.7944                                                            & 1.3321                                                              & \textbf{0.7463} \\
\multicolumn{1}{c|}{}                                                                         & mean     & 0.4909                                                            & 0.4865                                                            & 0.8448                                                              & \textbf{0.4605} \\ \hline
\multicolumn{1}{c|}{\multirow{4}{*}{IoU}}                                                     & airplane & 73.24                                                             & 74.33                                                             & 71.52                                                               & \textbf{74.53}  \\
\multicolumn{1}{c|}{}                                                                         & chair    & 78.81                                                             & 78.69                                                             & 74.42                                                               & \textbf{79.61}  \\
\multicolumn{1}{c|}{}                                                                         & table    & 73.75                                                             & \textbf{76.44}                                                    & 71.68                                                               & 75.85           \\
\multicolumn{1}{c|}{}                                                                         & mean     & 75.27                                                             & 76.49                                                             & 72.54                                                               & \textbf{76.66}  \\ \hline
\multicolumn{1}{c|}{\multirow{4}{*}{\begin{tabular}[c]{@{}c@{}}per-label\\ IoU\end{tabular}}} & airplane & 56.4                                                              & 57.7                                                              & 54.3                                                                & \textbf{67.8}   \\
\multicolumn{1}{c|}{}                                                                         & chair    & 60.3                                                              & \textbf{84.9}                                                     & 72.5                                                                & 81.5            \\
\multicolumn{1}{c|}{}                                                                         & table    & 78.5                                                              & 83.8                                                              & 76.0                                                                & \textbf{91.2}   \\
\multicolumn{1}{c|}{}                                                                         & mean     & 65.1                                                              & 75.5                                                              & 67.6                                                                & \textbf{80.2}   \\
\bottomrule
\end{tabular}
\caption{Quantitative results of ablation experiments. The CD values are multiplied by $10^{3}$; lower numbers are better. IoU values are multiplied by $10^2$; higher numbers are better. }
\label{tab:ablation}%
\end{table}

\subsection{Structured 3D Shape Autoencoding}
\label{sec:ResultsEval}

\begin{figure}
    \centering
    \includegraphics[width=0.99\linewidth]{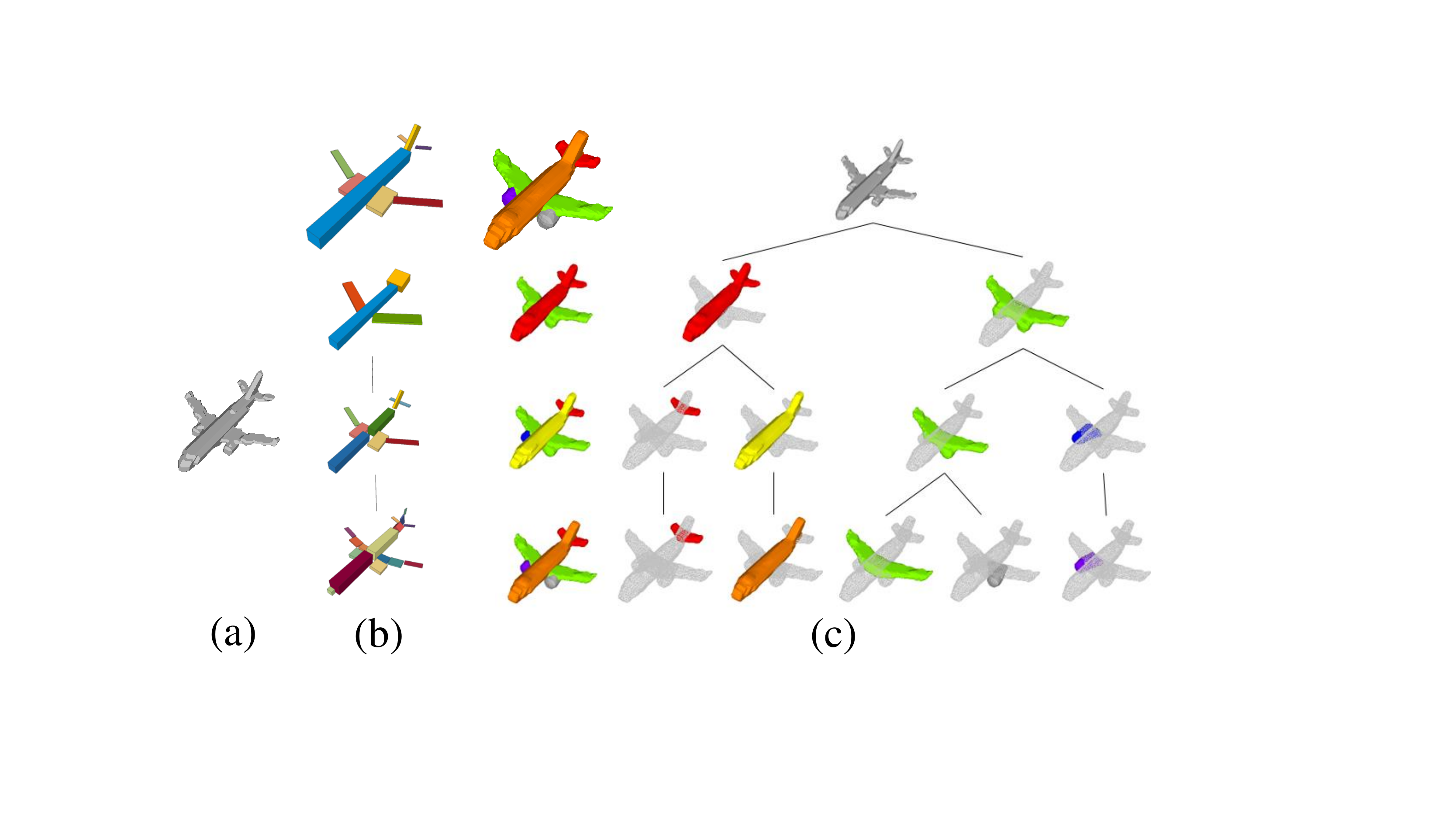}
    \caption{Qualitative comparison of structure hierarchies for a 3D input (a). HA~\cite{sun2019learning} predicts a 3-level hierarchy (b), with the top row as final output. Our predicted hierarchy is in (c): left column is the output at each level and the top one is the output of level 3. }
    \label{fig:3daeHiera}
\end{figure}

\begin{figure*}
    \centering
    \includegraphics[width=0.96\textwidth]{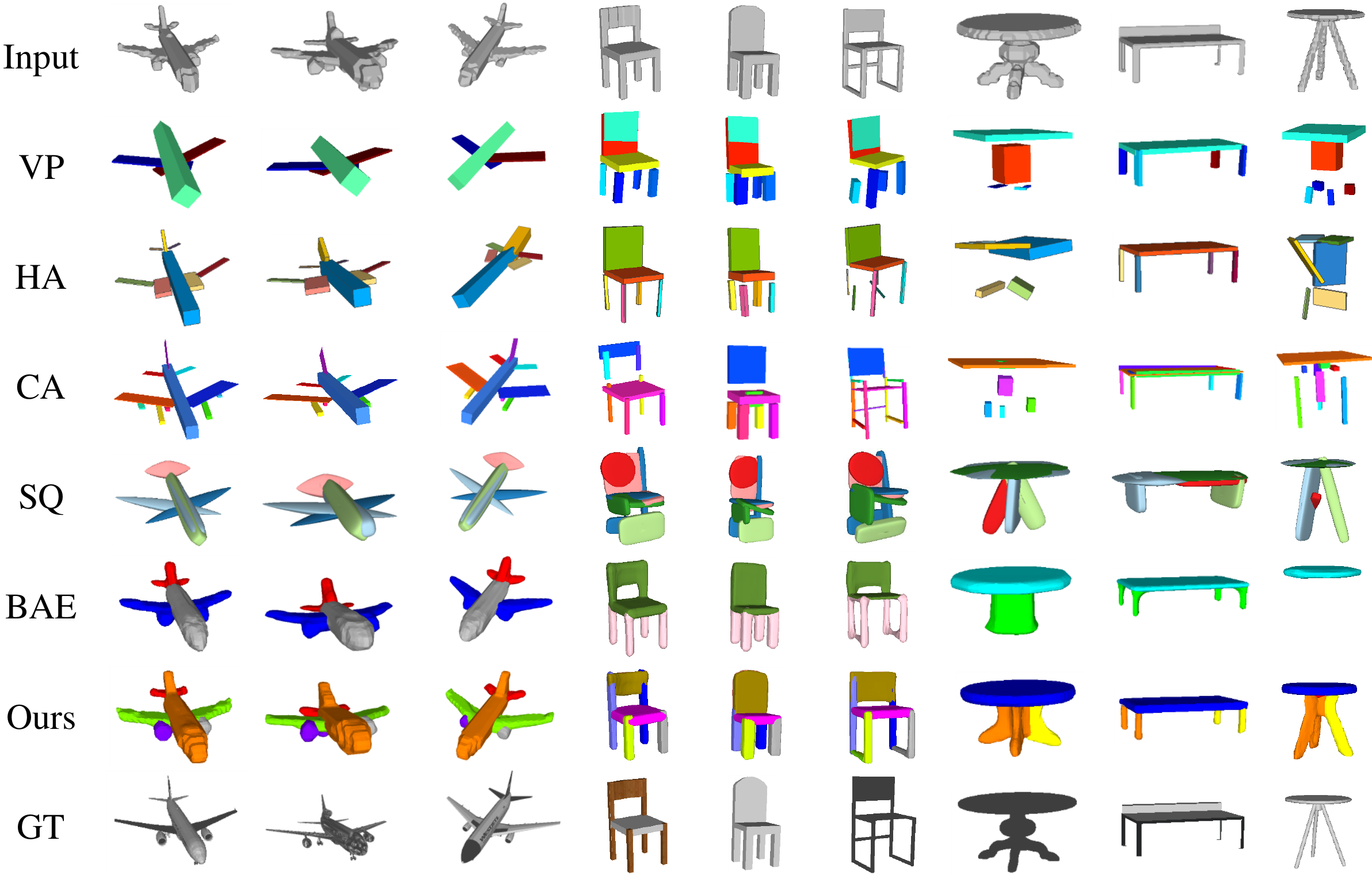}
    \caption{Visual comparison on structure learning methods VP~\cite{tulsiani2017learning}, HA~\cite{sun2019learning}, CA~\cite{yang2021unsupervised}, SQ~\cite{paschalidou2019superquadrics}, and BAE\cite{chen2019bae}. Distinct colors indicate primitive decomposition by the methods. The colored parts visualize segmentation consistency across different shapes in the same category.}
    \label{fig:comp}
\end{figure*}

\begin{table*}
 \setlength\tabcolsep{4.0pt}
  \centering
    \begin{tabular}{c||cccc||cccc||cccc}
    \toprule
    \multirow{2}[4]{*}{} & \multicolumn{4}{c||}{Chamfer Distance (CD)} & \multicolumn{4}{c||}{Intersection over Union (IoU)} & \multicolumn{4}{c}{Light Field Distance (LFD)} \\
\cmidrule{2-13}          & airplane & chair & table & Mean  & airplane & chair & table & Mean  & airplane & chair & table & Mean \\
    \midrule
    \midrule
    VP~\cite{tulsiani2017learning}    & 0.4587 & 0.7989 & 1.1561 & 0.8046 & 72.68 & 73.82 & 68.61 & 71.70 & 8698.83 & 4936.30 & 5101.14 & 6245.42 \\
    HA~\cite{sun2019learning}    & 0.2802 & 0.7199 & 1.0034 & 0.6678 & \textbf{77.35} & 77.56 & 72.25 & 75.72 & 7347.78 & 4459.68 & 4557.48 & 5454.98 \\
    CA~\cite{yang2021unsupervised}    & 0.3609 & 0.6682 & 0.9952 & 0.6748 & 69.25 & 72.32 & 68.46 & 70.01 & 6997.06 & 4600.93 & 4749.04 & 5449.01 \\
    SQ~\cite{paschalidou2019superquadrics}    & 0.3601 & 1.1654 & 1.3125 & 0.9460 & 67.26 & 69.38 & 69.94 & 68.86 & 7481.02 & 6745.72 & 6118.73 & 6781.82 \\
    BAE~\cite{chen2019bae}   & 0.4276 & 0.6945 & 1.2775 & 0.7999 & 69.34 & 72.29 & 62.57 & 68.07 & 6624.12 & 4139.83 & 4884.50 & 5216.15 \\
    Ours  & \textbf{0.2228} & \textbf{0.4125} & \textbf{0.7463} & \textbf{0.4605} & 74.53 & \textbf{79.61} & \textbf{75.85} & \textbf{76.66} & \textbf{5197.79} & \textbf{3410.60} & \textbf{3223.38} & \textbf{3943.92} \\
    \bottomrule
    \end{tabular}%
    \vspace{-2pt}
  \caption{Quantitative comparison of various structure learning methods. We report CD/LFD (lower is better) and IoU (higher is better). }
  \vspace{-3pt}
    \label{table:3DAE}%
\end{table*}

\begin{table}[htbp]
  \centering
    \begin{tabular}{c||c|c|c}
    \toprule
    Shape (\#parts) & airplane(4) & chair(4) & table(2) \\
    \midrule
    \begin{tabular}[c]{@{}c@{}}Segmented\\ parts\end{tabular} & \begin{tabular}[c]{@{}c@{}}body, tail,\\ wing, engine\end{tabular} & \begin{tabular}[c]{@{}c@{}}back, seat,\\ leg, arm\end{tabular} & \begin{tabular}[c]{@{}c@{}}top,\\ support\end{tabular}  \\
    \midrule
    \midrule
    VP~\cite{tulsiani2017learning}    & 37.6  & 64.7  & 62.1 \\
    HA~\cite{sun2019learning}    & 55.6  & 80.4  & 67.4 \\
    CA~\cite{yang2021unsupervised}    & 64.2  & \textbf{82.0}  & 89.2 \\
    SQ~\cite{paschalidou2019superquadrics}    & 48.9  & 65.6  & 77.7 \\
    BAE~\cite{chen2019bae}   & 61.1  & 65.5  & 87.0 \\
    Ours  & \textbf{67.8} & 81.5 & \textbf{91.2} \\
    \bottomrule
    \end{tabular}%
    \caption{Quantitative results of per-label IoU (higher is better).}    
    \vspace{-10pt}
  \label{tab:partiou}%
\end{table}%

Our method learns hierarchical shape structures without ground-truth structure. We hence compare with alternative self-supervised structure learning methods.
We set $3$ levels with a maximum of $8$ primitives for airplanes and tables. For chairs, we use $4$ levels with up to $16$ primitives. \rz{These maximum primitive count settings are} the same as SQ~\cite{paschalidou2019superquadrics} and BAE-Net~\cite{chen2019bae}. For training SQ and CA~\cite{yang2021unsupervised}, we use the default training data processing and training settings released by the authors. Results of VP~\cite{tulsiani2017learning} and HA~\cite{sun2019learning} are published by the author of HA using the default training settings (the maximum primitive count is $16$ for airplanes, $12$ for tables, and $32$ for chairs). The comparison is done with the shared portion of the test split by the various methods.

\vspace{-10pt}

\paragraph{Structure hierarchy.}
%
HA~\cite{sun2019learning} is a representative work of learning hierarchical shape structures in a self-supervised manner. Since there is no well-established evaluation metric or protocol available, we provide a qualitative comparison of the learned hierarchies in \cref{fig:3daeHiera}.
HA decomposes a part into smaller ones into deep levels and then selects a set of reasonable parts across various levels as the final output. In our method, the decomposition at each level is always a full reconstruction of the input model. As shown in \cref{fig:comp}, our decomposition results look more consistent and meaningful. HA learns shape hierarchies through abstracting shape parts as cuboids, which may overly decompose a geometrically complex part into many small cuboids. In contrast, our method models the geometric variations of shape parts with implicit fields, leading to more consistent structures.

\vspace{-5pt}

\paragraph{Structure reconstruction.}
We compare with more structure learning methods. Since most of the structure learning methods do not produce hierarchy, we compare them with the finest level of our recursive implicit fields. In the quantitative comparison in \cref{table:3DAE}, our method outperforms all alternatives in CD, IoU and LFD, especially on tables and chairs with rich structures. \cref{fig:comp} provides a visual comparison. VP~\cite{tulsiani2017learning} and SQ~\cite{paschalidou2019superquadrics} abstract an object into sets of cuboids or superquadrics, respectively, causing the loss of fine details and the meaningfulness of the parts, especially for chairs. HA~\cite{sun2019learning} learns better structures with its hierarchical cuboid abstraction, but is still unable to handle the shape variations on, e.g., the round table and the four-legged table. CA~\cite{yang2021unsupervised} can generate shape structures with more details, but shares the same drawback as HA. BAE-Net~\cite{chen2019bae} preserves fine details of the objects but only captures coarse parts. In contrast, RIM-Net is able to learn fine-grained parts with accurate reconstruction, under the same setting of maximum part count. For example, BAE-Net fails to segment the individual legs and cannot separate the back from the seat of chairs while our method succeeds.

\vspace{-10pt}

\paragraph{Structure co-segmentation.}
Next, we evaluate structure learning from the segmentation point of view, measuring the consistency and meaningfulness of the segments. In particular, we transfer the segmentations onto point clouds as per-point part label and compute the mIoU to evaluate co-segmentation quality, following~\cite{chen2019bae}. Since different methods output different numbers of parts for the same category, we take the ground-truth dataset of~\cite{yi2016scalable} and conduct voting-based label snapping as in~\cite{chen2019bae}, for a fair comparison. The ``leg'' and ``support'' labels are merged as ``support'' for the table category. From \cref{tab:partiou}, our method is superior to other unsupervised structure learning methods.

\subsection{Single-View Reconstruction}

Finally, we apply RIM-Net to infer hierarchical shape structures in 3D from a single image input.

\begin{figure}
    \centering
    \includegraphics[width=0.99\linewidth]{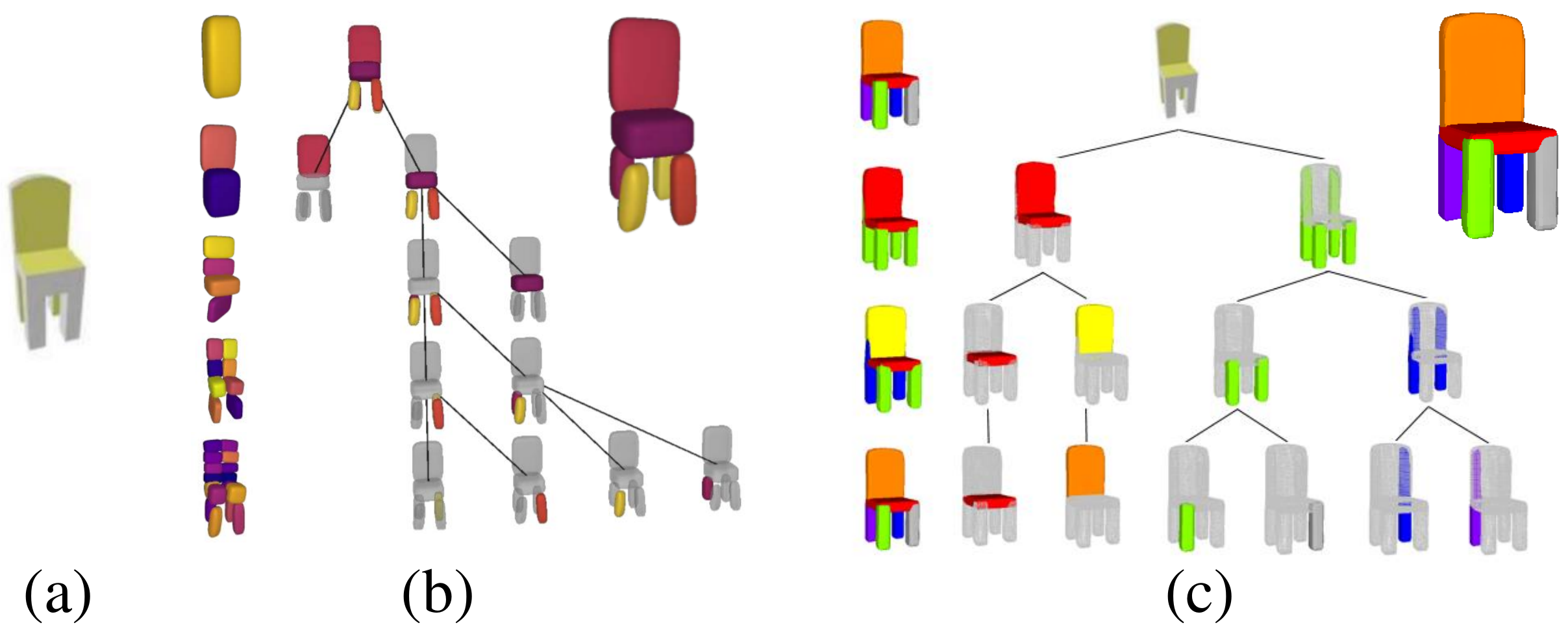}
    \caption{Qualitative comparison of single-view reconstruction of hierarchical structures. Given an input image (a), ~\cite{paschalidou2020learning} predicts a top-down hierarchy (b). The left column is the output of each level. The final output is in the top-right corner of the hierarchy. (c) is our predicted structure hierarchy. The output of each level is shown to the left, and the output of level 3 in the top-right corner. }
    \label{fig:svrHiera}
\end{figure}

\vspace{-10pt}

\paragraph{Structure hierarchy.}
On unsupervised single-view reconstruction of hierarchical structures, the work of Paschalidou et al.~\cite{paschalidou2020learning} is the most relevant. Their work predicts structure hierarchies based on spatial positions: A part is decomposed in three ways (up-down, front-back, and left-right) without accounting for the meaningfulness of parts. We visually compare the hierarchies of the two methods in \cref{fig:svrHiera}. The shape expression ability of superquadrics of \cite{paschalidou2020learning} is limited. Moreover, their decomposition could cause over-segmentation. For example, the back of the chair is segmented at each level, which is obviously unnecessary. In our hierarchy, each level constitutes a good approximation of the input, and the segments at each level are semantically meaningful. More importantly, their method cannot produce consistent segmentation for shapes of the same category, due to spatial position based decomposition. 

\vspace{-10pt}

\paragraph{Implicit fields reconstruction.}
We compare to alternative methods, IM-Net~\cite{chen2019learning} and BSP-Net~\cite{chen2020bsp}, which also employ implicit fields, to demonstrate how our key designs, the hierarchy and the per-point Gaussians, improve reconstruction quality. Different from our method, IM-Net outputs a single implicit field for the holistic shape while BSP-Net outputs a set of convexes together forming the field.

For this experiment, we use the same dataset as IM-Net. It contains five representative categories from ShapeNet~\cite{chang2015shapenet} with rendered views by 3D-R2N2~\cite{choy20163d}, i.e., airplanes (4,045), cars (7,497), chairs (6,778), rifles (2,373) and tables (8,509). For all methods, we train a separate model for each category. The train/test split we use is the same as~\cite{chen2020bsp}.
For the training of implicit field SVR, we adopt the training scheme in~\cite{chen2019learning,chen2020bsp}. We first pretrain a 3D auto-encoder. We then supervise the feature reconstruction quality of the image encoder module through measuring the mean square error (MSE) between the features extracted for the input image and the pre-trained features of the 3D auto-encoder.

\begin{figure}
    \centering
    \includegraphics[width=0.99\linewidth]{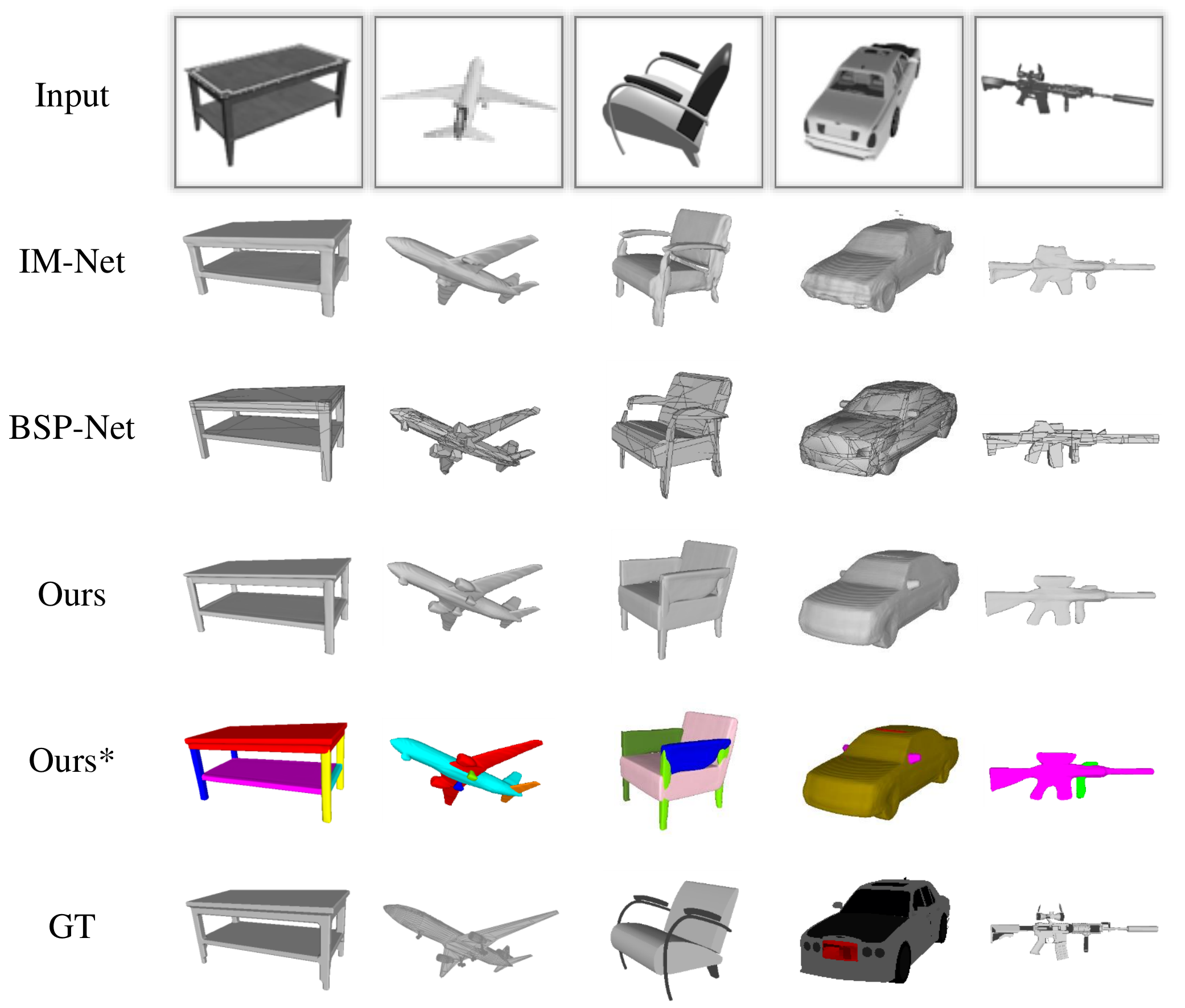}
    \caption{Single view reconstruction results of various methods. We show all the reconstructed results in gray color. ``Ours*'' means our results with colored parts reflecting the inferred structures. ``GT'' denotes ground-truth objects.}
    \label{fig:svr}
\end{figure}

\begin{table}[t!]
\scriptsize
  \centering
  \setlength\tabcolsep{2.0pt}
    \begin{tabular}{c||ccc|ccc}
    \toprule
    \multirow{2}[4]{*}{Category} & \multicolumn{3}{c|}{Chamfer Distance (CD)} & \multicolumn{3}{c}{Light Field Distance (LFD)} \\
\cmidrule{2-7}          & IM-Net~\cite{chen2019learning}  & BSP-Net~\cite{chen2020bsp} & Ours  & IM-Net~\cite{chen2019learning} & BSP-Net~\cite{chen2020bsp} & Ours \\
    \midrule
    \midrule
    airplane & 0.4041 & 0.4716 & \textbf{0.3993} & 5725.02 & 5397.83 & \textbf{5252.53} \\
    car   & 0.6833 & 0.6262 & \textbf{0.4699} & 2788.74 & 2834.14 & \textbf{2686.94} \\
    chair & 0.8799 & 0.7472 & \textbf{0.7446} & 3499.15 & \textbf{3371.78} & 3600.60 \\
    rifle & 0.4439 & 0.5708 & \textbf{0.4095} & 6698.26 & 8834.54 & \textbf{6313.92} \\
    table & \textbf{0.8762} & 0.9807 & 1.1223 & \textbf{3232.52} & 3251.41 & 3511.95 \\
    mean  & 0.6575 & 0.6793 & \textbf{0.6291} & 4388.74 & 4737.94 & \textbf{4273.19} \\
    \bottomrule
    \end{tabular}%
  \caption{Quantitative comparison of single view reconstruction.}
  \label{tab:svr}%
\end{table}%

As reported in~\cref{tab:svr}, our method outperforms the two methods in CD and LFD. \cref{fig:svr} shows qualitative results. As demonstrated in the visual results, our method achieves comparable reconstruction quality with the state of the arts, while additionally producing hierarchical structures without any supervision. The inferred segmentations by our method are shown in colors in the row of ``Ours*''.



\section{Conclusion, limitation, and future work}
\label{sec:Conclusion}

We have introduced RIM-Net, a learning-based hierarchical framework that generates shape with implicit primitives without requiring any part-level labels for training. The per-point Gaussian as local point distribution is inserted into the point label prediction process, which effectively enhances the part decomposition and detail generation. A constraint of our method is that the hierarchy topologies are arbitrary and there is no standard to measure them.

An interesting future study will be to explore whether our model can be plugged into various learning-based prediction methods to generate hierarchical primitives for objects. Another future improvement is to use more constraints to construct the object in a diverse hierarchy, like various part prior, symmetry of shape.

\section*{Acknowledgments}
\ncj{We thank the anonymous reviewers for their valuable comments, and Qimin Chen from SFU for his earlier help on the project.
This work was supported in part by the NSFC (62132021, 62002375, 62002376, 62102435, 61902419), National Key Research and Development Program of China (2018AAA0102200), NSERC (611370), and gift funds from Adobe, Autodesk, and Google.}






{\small
\bibliographystyle{ieee_fullname}
\bibliography{egbib}

\begin{thebibliography}{10}\itemsep=-1pt

\bibitem{carlson1999}
L. Carlson-Radvansky, E. Covey, and K. Lattanzi.
\newblock {``What''} effects on ``where'': Functional influence on spatial
  relations.
\newblock {\em Psychological Science}, 10(6):519--521, 1999.

\bibitem{chang2015shapenet}
Angel~X Chang, Thomas Funkhouser, Leonidas Guibas, Pat Hanrahan, Qixing Huang,
  Zimo Li, Silvio Savarese, Manolis Savva, Shuran Song, Hao Su, et~al.
\newblock Shapenet: An information-rich 3d model repository.
\newblock {\em arXiv preprint arXiv:1512.03012}, 2015.

\bibitem{chen2020bsp}
Zhiqin Chen, Andrea Tagliasacchi, and Hao Zhang.
\newblock Bsp-net: Generating compact meshes via binary space partitioning.
\newblock In {\em Proceedings of the IEEE/CVF Conference on Computer Vision and
  Pattern Recognition}, pages 45--54, 2020.

\bibitem{chen2019bae}
Zhiqin Chen, Kangxue Yin, Matthew Fisher, Siddhartha Chaudhuri, and Hao Zhang.
\newblock {BAE-Net}: Branched autoencoder for shape co-segmentation.
\newblock In {\em Proceedings of the IEEE/CVF International Conference on
  Computer Vision}, pages 8490--8499, 2019.

\bibitem{chen2019learning}
Zhiqin Chen and Hao Zhang.
\newblock Learning implicit fields for generative shape modeling.
\newblock In {\em Proceedings of the IEEE/CVF Conference on Computer Vision and
  Pattern Recognition}, pages 5939--5948, 2019.

\bibitem{choy20163d}
Christopher~B Choy, Danfei Xu, JunYoung Gwak, Kevin Chen, and Silvio Savarese.
\newblock 3d-r2n2: A unified approach for single and multi-view 3d object
  reconstruction.
\newblock In {\em European conference on computer vision}, pages 628--644.
  Springer, 2016.

\bibitem{genova2019learning}
Kyle Genova, Forrester Cole, Daniel Vlasic, Aaron Sarna, William~T Freeman, and
  Thomas Funkhouser.
\newblock Learning shape templates with structured implicit functions.
\newblock In {\em Proceedings of the IEEE/CVF International Conference on
  Computer Vision}, pages 7154--7164, 2019.

\bibitem{hoffman1984}
D.~D. Hoffman and W.~A. Richards.
\newblock Parts of recognition.
\newblock {\em Cognition}, pages 65--96, 1984.

\bibitem{kalogerakis20173d}
Evangelos Kalogerakis, Melinos Averkiou, Subhransu Maji, and Siddhartha
  Chaudhuri.
\newblock 3d shape segmentation with projective convolutional networks.
\newblock In {\em proceedings of the IEEE conference on computer vision and
  pattern recognition}, pages 3779--3788, 2017.

\bibitem{li2017grass}
Jun Li, Kai Xu, Siddhartha Chaudhuri, Ersin Yumer, Hao Zhang, and Leonidas
  Guibas.
\newblock Grass: Generative recursive autoencoders for shape structures.
\newblock {\em ACM Transactions on Graphics (TOG)}, 36(4):1--14, 2017.

\bibitem{maturana2015voxnet}
Daniel Maturana and Sebastian Scherer.
\newblock Voxnet: A 3d convolutional neural network for real-time object
  recognition.
\newblock In {\em 2015 IEEE/RSJ International Conference on Intelligent Robots
  and Systems (IROS)}, pages 922--928. IEEE, 2015.

\bibitem{mescheder2019occupancy}
Lars Mescheder, Michael Oechsle, Michael Niemeyer, Sebastian Nowozin, and
  Andreas Geiger.
\newblock Occupancy networks: Learning 3d reconstruction in function space.
\newblock In {\em Proceedings of the IEEE/CVF Conference on Computer Vision and
  Pattern Recognition}, pages 4460--4470, 2019.

\bibitem{mo2019structurenet}
Kaichun Mo, Paul Guerrero, Li Yi, Hao Su, Peter Wonka, Niloy Mitra, and
  Leonidas~J Guibas.
\newblock Structurenet: Hierarchical graph networks for 3d shape generation.
\newblock {\em ACM Transactions on Graphics (TOG)}, 39(1):1--19, 2019.

\bibitem{muralikrishnan2018tags2parts}
Sanjeev Muralikrishnan, Vladimir~G Kim, and Siddhartha Chaudhuri.
\newblock Tags2parts: Discovering semantic regions from shape tags.
\newblock In {\em Proceedings of the IEEE Conference on Computer Vision and
  Pattern Recognition}, pages 2926--2935, 2018.

\bibitem{niu2018im2struct}
Chengjie Niu, Jun Li, and Kai Xu.
\newblock Im2struct: Recovering 3d shape structure from a single rgb image.
\newblock In {\em Proceedings of the IEEE conference on computer vision and
  pattern recognition}, pages 4521--4529, 2018.

\bibitem{palmer1977}
Stephen~E. Palmer.
\newblock Hierarchical structure in perceptual representation.
\newblock {\em Cognitive Psychology}, 9(4):441--474, 1977.

\bibitem{park2019deepsdf}
Jeong~Joon Park, Peter Florence, Julian Straub, Richard Newcombe, and Steven
  Lovegrove.
\newblock {DeepSDF}: Learning continuous signed distance functions for shape
  representation.
\newblock In {\em Proceedings of the IEEE Conference on Computer Vision and
  Pattern Recognition}, pages 165--174, 2019.

\bibitem{paschalidou2020learning}
Despoina Paschalidou, Luc~Van Gool, and Andreas Geiger.
\newblock Learning unsupervised hierarchical part decomposition of 3d objects
  from a single {RGB} image.
\newblock In {\em Proceedings of the IEEE/CVF Conference on Computer Vision and
  Pattern Recognition}, pages 1060--1070, 2020.

\bibitem{paschalidou2019superquadrics}
Despoina Paschalidou, Ali~Osman Ulusoy, and Andreas Geiger.
\newblock Superquadrics revisited: Learning 3d shape parsing beyond cuboids.
\newblock In {\em Proceedings of the IEEE/CVF Conference on Computer Vision and
  Pattern Recognition}, pages 10344--10353, 2019.

\bibitem{qi2017pointnet}
Charles~R Qi, Hao Su, Kaichun Mo, and Leonidas~J Guibas.
\newblock Pointnet: Deep learning on point sets for 3d classification and
  segmentation.
\newblock In {\em Proceedings of the IEEE conference on computer vision and
  pattern recognition}, pages 652--660, 2017.

\bibitem{qi2017pointnet++}
Charles~R Qi, Li Yi, Hao Su, and Leonidas~J Guibas.
\newblock Pointnet++: Deep hierarchical feature learning on point sets in a
  metric space.
\newblock In {\em Advances in neural information processing systems}, pages
  5105--5114, 2017.

\bibitem{sun2019learning}
Chun-Yu Sun, Qian-Fang Zou, Xin Tong, and Yang Liu.
\newblock Learning adaptive hierarchical cuboid abstractions of 3d shape
  collections.
\newblock {\em ACM Transactions on Graphics (TOG)}, 38(6):1--13, 2019.

\bibitem{thompson1992}
D.~W. Thompson.
\newblock {\em On Growth and Form}.
\newblock Dover, 1992.

\bibitem{tulsiani2017learning}
Shubham Tulsiani, Hao Su, Leonidas~J Guibas, Alexei~A Efros, and Jitendra
  Malik.
\newblock Learning shape abstractions by assembling volumetric primitives.
\newblock In {\em Proceedings of the IEEE Conference on Computer Vision and
  Pattern Recognition}, pages 2635--2643, 2017.

\bibitem{conshier_sig13}
Oliver van Kaick, Kai Xu, Hao Zhang, Yanzhen Wang, Shuyang Sun, Ariel Shamir,
  and Daniel Cohen-Or.
\newblock Co-hierarchical analysis of shape structures.
\newblock {\em ACM Transactions on Graphics (Special Issue of SIGGRAPH)},
  32(4):Article 69, 2013.

\bibitem{yang2021unsupervised}
Kaizhi Yang and Xuejin Chen.
\newblock Unsupervised learning for cuboid shape abstraction via joint
  segmentation from point clouds.
\newblock {\em ACM Transactions on Graphics (TOG)}, 40(152):1--11, 2021.

\bibitem{yi2016scalable}
Li Yi, Vladimir~G Kim, Duygu Ceylan, I-Chao Shen, Mengyan Yan, Hao Su, Cewu Lu,
  Qixing Huang, Alla Sheffer, and Leonidas Guibas.
\newblock A scalable active framework for region annotation in 3d shape
  collections.
\newblock {\em ACM Transactions on Graphics (ToG)}, 35(6):1--12, 2016.

\bibitem{yu2019partnet}
Fenggen Yu, Kun Liu, Yan Zhang, Chenyang Zhu, and Kai Xu.
\newblock Partnet: A recursive part decomposition network for fine-grained and
  hierarchical shape segmentation.
\newblock In {\em Proceedings of the IEEE/CVF Conference on Computer Vision and
  Pattern Recognition}, pages 9491--9500, 2019.

\bibitem{zhao20193d}
Yongheng Zhao, Tolga Birdal, Haowen Deng, and Federico Tombari.
\newblock 3d point capsule networks.
\newblock In {\em Proceedings of the IEEE/CVF Conference on Computer Vision and
  Pattern Recognition}, pages 1009--1018, 2019.

\bibitem{zhu2020adacoseg}
Chenyang Zhu, Kai Xu, Siddhartha Chaudhuri, Li Yi, Leonidas~J Guibas, and Hao
  Zhang.
\newblock Adacoseg: Adaptive shape co-segmentation with group consistency loss.
\newblock In {\em Proceedings of the IEEE/CVF Conference on Computer Vision and
  Pattern Recognition}, pages 8543--8552, 2020.

\bibitem{zou20173d}
Chuhang Zou, Ersin Yumer, Jimei Yang, Duygu Ceylan, and Derek Hoiem.
\newblock 3d-prnn: Generating shape primitives with recurrent neural networks.
\newblock In {\em Proceedings of the IEEE International Conference on Computer
  Vision}, pages 900--909, 2017.

\end{thebibliography}
}

\end{document}